\documentclass[10pt]{article} 
\usepackage[preprint]{tmlr}

\usepackage{amsmath,amsfonts,bm}

\def\eqref#1{equation~\ref{#1}}

\def\1{\bm{1}}

\DeclareMathAlphabet{\mathsfit}{\encodingdefault}{\sfdefault}{m}{sl}
\SetMathAlphabet{\mathsfit}{bold}{\encodingdefault}{\sfdefault}{bx}{n}

\usepackage{hyperref}
\usepackage{url}
\usepackage{graphicx}
\usepackage{amsmath}
\usepackage{amssymb}
\usepackage{algorithm}
\usepackage{algpseudocode}
\usepackage{booktabs}
\usepackage{comment}
\usepackage[most]{tcolorbox}

\title{Skeleton-and-Strategy Prompting: Training-Free\\Negation Understanding for Vision-Language Models}

\author{\name Yuliang Cai \email caiyulia@usc.edu \\
      \addr Viterbi School of Engineering\\
      University of Southern California
      \AND
      \name Mohammad Rostami
      \email rostamii@amazon.com\\
      \addr Amazon Inc.
      \AND
      \name Jesse Thomason \email jessetho@usc.edu \\
      \addr Viterbi School of Engineering\\
      University of Southern California
      }

\def\month{MM}  
\def\year{YYYY} 
\def\openreview{\url{https://openreview.net/forum?id=XXXX}} 

\begin{document}

\maketitle

\begin{abstract}
Despite the strong performance of Vision-Language Models (VLMs) on a wide range of visual question answering (VQA) tasks, these models consistently struggle to understand negation and produce incorrect answers when questions involve negated clauses. 
To address this limitation, we propose Skeleton-and-Strategy Prompting (\textbf{SSP}), a training-free, in-context learning method that improves VLM negation understanding capabilities without any parameter updates. 
Given a negation question, our method first abstracts the underlying question structure into a skeleton, retrieves a small set of same-skeleton questions from a lightweight question pool, then prompts the VLM to analyze their shared negation pattern and synthesize a single-sentence answering strategy. The skeleton and strategy are prepended to the test sample to guide the model correctly tackle the negation problems. Experiments on multiple negation VQA benchmarks show that SSP achieves state-of-the-art performance on negation-focused VQA tasks while remaining computationally efficient.
\end{abstract}

\section{Introduction}

Vision-language models (VLMs) have shown strong performance across a wide range of visual question answering (VQA) tasks, demonstrating their ability to reason about complex visual scenes from natural language instructions. However, these models lack the concrete understanding of \textbf{negation}, which is a fundamental component of the human language, enabling the expression of absence, exclusion, and logical opposition in everyday communication. Visual question answering tasks can include negation, for example \textit{"Is there a dog in the image but no cat?"} or \textit{"Which objects are not on the table?"} 
When a VLM fails to understand negation, it tends to respond as if the negation were absent, producing incorrect answers that undermine the model's reliability.

\begin{figure}[ht]
    \centering
    \includegraphics[width=\linewidth]{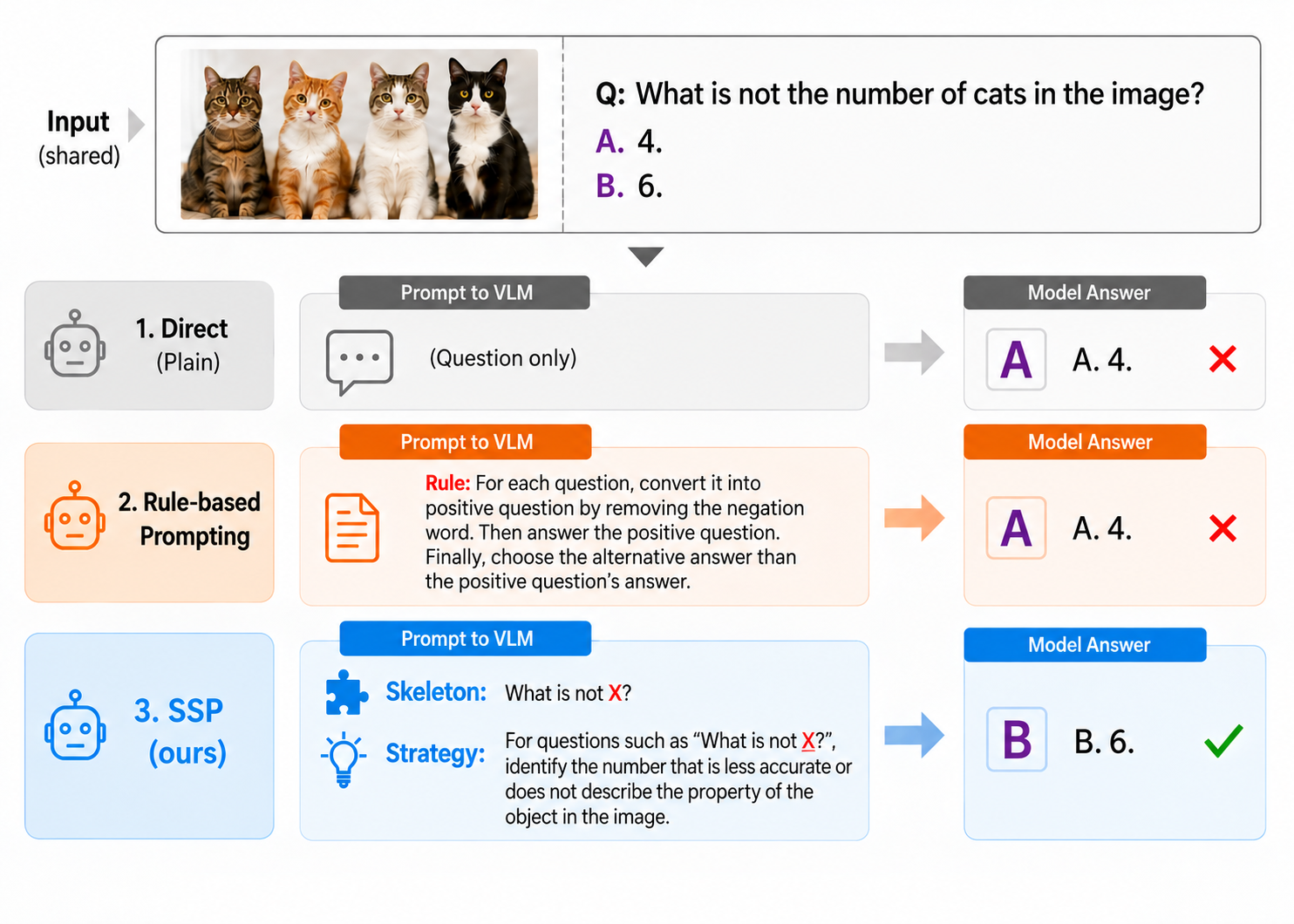}
    \caption{\textbf{Comparison of prompting methods under the negation scenario.}
Given the same negation question to VLMs, the plain prompt and rule-based prompt both produce the answers which align with the visual content.
Our \textbf{Skeleton-and-Strategy Prompting (SSP)} extracts the question skeleton and applies a negation-aware strategy, guiding the model to choose the option that differs from the visual content, which is the correct answer for the negation question.
}
    \label{fig:open_demonstration}
\end{figure}

Existing studies in the negation understanding of VLMs have only primarily focused on CLIP-based models~\citep{radford2021learningtransferablevisualmodels} that use image-text alignment tasks where the objective is to measure similarity between global image and text representations. In this setting, handling negation is typically framed as distinguishing between matching and non-matching pairs, without requiring deeper compositional or logical reasoning. These methods rely on implicit learning signals within the contrastive framework and do not explicitly model the semantics of negation, limiting their ability to capture fine-grained logical distinctions beyond surface-level mismatches. Moreover, most existing approaches adopt full-model fine-tuning to adapt CLIP to negation-aware tasks~\citep{cai2025tngcliptrainingtimenegationdatageneration,park2025knownobetterdatadriven,singh2024learnnosayyes,yuksekgonul2023visionlanguagemodelsbehavelike}, which can be computationally expensive and inefficient when finetuning large pretrained models.

To address the lack of negation understanding methods in the reasoning task, we propose an efficient, training-free method for enhancing negation understanding in vision-language models for visual question answering, a setting that requires reasoning not about what is present but about what is absent (Figure~\ref{fig:open_demonstration}). We show that negation VQA can be treated as a structure-sensitive reasoning problem. In many cases, a pretrained VLM already has the knowledge needed to answer the question, such as recognizing the objects, attributes, or relations in the image. However, the model may still fail because it does not correctly interpret the negated sentence pattern. Thus, we introduce Skeleton-and-Strategy Prompting (\textbf{SSP}): given a negation question, we first abstract its underlying structure into a skeleton (e.g., \textit{"What is not [X]?"}), retrieve a small set of same-skeleton questions from a lightweight question pool, and prompt the VLM to analyze their shared negation pattern and synthesize a single-sentence answering strategy, which is a concise instruction that tells the VLM how to interpret the negaiton structure and select the correct answer. 
Prepending this skeleton and strategy to the prompt containing the test sample helps the model correctly handle the negation.
Because SSP caches the skeleton and pool, as inference proceeds SSP is able to re-use previously calculated skeletons when new questions match them, reducing computational overhead.
Experimental results demonstrate that \textbf{SSP} improves negation understanding in VQA tasks while maintaining minimal overhead.

\section{Related Works}

Our work lies at the intersection of two research directions: in-context learning for visual question answering and negation understanding in vision-language models. Multimodal in-context learning demonstrates how VLMs can adapt at inference time, but existing approaches typically rely on instance-level demonstrations rather than reusable abstractions of the underlying reasoning structure. Meanwhile, prior work on visual negation has shown that VLMs consistently struggle to interpret negated inputs, but has focused primarily on image–text alignment and training-based adaptation. These limitations motivate \textbf{SSP}, which provides training-free guidance for negation reasoning by converting shared question structures into reusable skeleton–strategy prompts.

\subsection{In-Context Learning for Visual Question Answering}

In-context learning (ICL) enables large models to adapt to new tasks through demonstrations provided at inference time, eliminating the need for parameter updates~\citep{brown2020languagemodelsfewshotlearners}. Building on this paradigm, multimodal models such as Frozen~\citep{tsimpoukelli2021multimodalfewshotlearningfrozen} and Flamingo~\citep{alayrac2022flamingovisuallanguagemodel} extend few-shot in-context learning to vision-language tasks, conditioning on interleaved image-text exemplars to guide reasoning at inference time without any gradient updates.

Beyond general-purpose multimodal ICL, a separate line of work has targeted training-free prompting strategies specifically for VQA. PICa~\citep{yang2022empiricalstudygpt3fewshot} showed that large language models can answer visual questions using image captions and a small set of demonstrations. PNP-VQA~\citep{tiong2023plugandplayvqazeroshotvqa} further leveraged caption generation and language-model reasoning to perform zero-shot VQA. Subsequent works explored prompt engineering, chain-of-thought reasoning, and multimodal rationale generation to improve visual reasoning~\citep{kojima2023largelanguagemodelszeroshot,zhang2024multimodalchainofthoughtreasoninglanguage,shao2024visualcotadvancingmultimodal,guo2023imagestextualpromptszeroshot}. More recent studies examined multimodal demonstration selection and evaluation, highlighting the importance of exemplar quality and retrieval strategies in multimodal ICL~\citep{zhao2024mmiclempoweringvisionlanguagemodel,zong2025vliclbenchdevildetails}. 

Despite these advances, existing approaches primarily transfer knowledge through explicit examples, requiring the model to infer the underlying reasoning pattern from retrieved demonstrations. This limitation is particularly significant for negation VQA, where success depends on correctly interpreting linguistic structures such as exclusion, absence, or contradiction.
\textbf{SSP} addresses this gap by explicitly extracting the shared negation-aware structure of related questions and converting it into a reusable reasoning strategy, thereby removing the need for the VLM to infer the reasoning pattern solely from instance-level demonstrations.

\subsection{Negation Understanding for VLM}

Negation is a fundamental linguistic phenomenon that has long posed challenges for natural language processing systems. From the perspective of language-only models, existing work has demonstrated that pretrained language models fail to distinguish between negated and non-negated queries~\citep{kassner-schutze-2020-negated,ettinger2020bertnotlessonsnew}.
LLMs have also been evaluated systematically via negation-focused benchmarks~\citep{truong-etal-2023-language,garcíaferrero2023datasetlargenegationbenchmark}, showing that LLMs exhibit several limitations to the presence of negation, an inability to capture the lexical semantics of negation, and a failure to reason under negation.

On the other hand, the study of negation in VLMs is mainly focused on CLIP-based model~\citep{radford2021learningtransferablevisualmodels}. For example, \citeauthor{quantmeyer2024doesclipprocessnegation} conduct experiments and visualize where and how the CLIP model processes negation information in each layer. To encourage processing negation semantics, other methods adopt LLMs to generate negation captions based on existing image–text pair datasets to fine-tune CLIP for negation understanding~\citep{alhamoud2025visionlanguagemodelsunderstandnegation,singh2024learnnosayyes,park2025knownobetterdatadriven,yuksekgonul2023visionlanguagemodelsbehavelike,cai2025tngcliptrainingtimenegationdatageneration}. 
However, the under-performance of negation is not only constrained in the CLIP-based model for image-text matching tasks, but widely exists in tasks which require more complicated reasoning, such as visual question answering. One study found that, in visual question answering with multiple choices, simply turning the question into a negated format (e.g. \textit{"What is in front of the door?"} to \textit{"What is not in front of the door?"}) caused severe drops in state-of-the-art VLM accuracy~\citep{zhang2025negvqavisionlanguagemodels}. 

We target improving the negation understanding for the visual question answering task.
Instead of performing full-model fine-tuning like the methods mentioned above, we propose a training-free in-context learning method, which leads to temporal and computational efficiency.

\begin{figure*}[t]
    \centering
    \includegraphics[width=\linewidth]{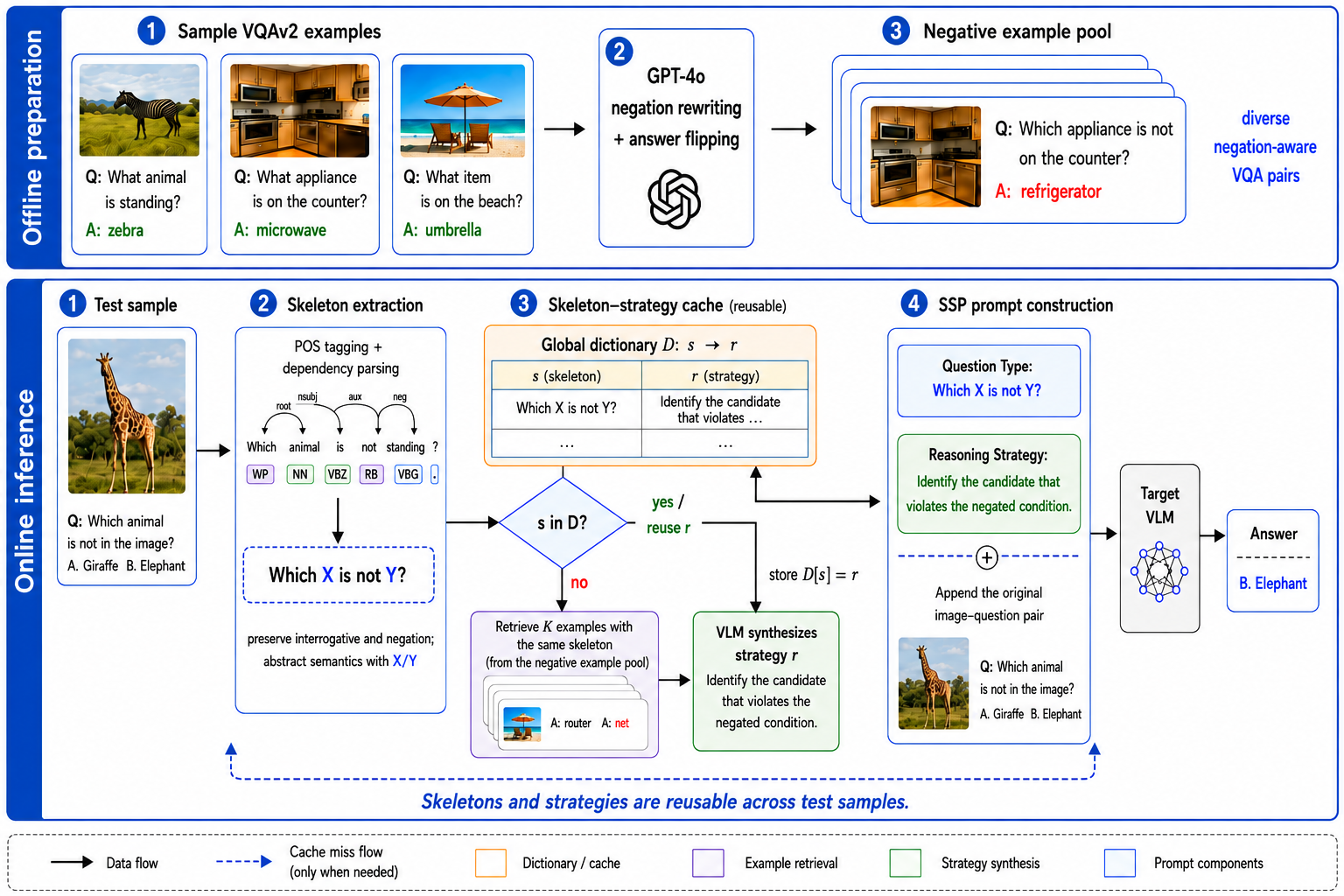}
    \vspace{-12pt}
    \caption{\textbf{Overview of Skeleton and Strategy Prompting (SSP).} SSP first builds a negative example pool offline (top). 
    At inference time, SSP extracts the question skeleton, retrieves or synthesizes a reusable reasoning strategy, and constructs the final prompt for the target VLM (bottom).
    }
    \label{fig:SSPpipeline}
\end{figure*}

\section{Methodology}
We propose \textbf{Skeleton and Strategy Prompting (SSP)}, a training-free, inference-time framework for negation VQA (Figure~\ref{fig:SSPpipeline}). \textbf{SSP} is motivated by the observation that questions sharing the same negation-aware format usually require similar reasoning procedures. Instead of retrieving demonstrations for every test instance, \textbf{SSP} abstracts each question into a reusable skeleton and associates it with a reasoning strategy. In particular, \textbf{SSP} maintains a global example pool, $\mathcal{D}: s \mapsto  r,$ where $s$ denotes the extracted skeleton, and $r$ denotes the corresponding reasoning strategy. Given a test question, \textbf{SSP} first identifies its question type. If this type has already been stored in $\mathcal{D}$, \textbf{SSP} directly reuses the cached skeleton and strategy. Otherwise, \textbf{SSP} extracts a new skeleton using a rule-based procedure and generates a new strategy using negative VQA samples from the example pool.

\subsection{Example Pool Construction}
To support the generation of question-specific strategies in \textbf{SSP}, we construct an example pool from an external VQA source. First, we randomly sample $K$ visual question answering examples from the VQA-v2 dataset~\citep{goyal2017makingvvqamatter}. Each sampled example consists of an image, an original question, its corresponding pair of answer options, and the label for the correct answer. Since VQA-v2 mainly contains standard positive-form questions, we use ${\text{GPT-4o}}$~\citep{openai2024gpt4ocard} to transform each sampled question into a negation-format question while preserving the original visual content and answer space. During this conversion, GPT-4o rewrites the question into a negative format, involving  absence, exclusion, or incorrect attributes. We then swap the original correct answer label to the alternative among the binary pair from VQA-v2 to obtain the target answer for the negated question. In this way, each example in the pool contains a negative question-answer pair derived from an existing VQA instance. The resulting example pool provides diverse negation patterns and visual reasoning cases, which are later used by \textbf{SSP} to synthesize strategies for unseen test questions. This construction allows \textbf{SSP} to leverage broad visual-language knowledge from VQA-v2 without requiring additional model training or task-specific fine-tuning. 

\subsection{Skeleton Extraction}
\label{Skeleton Extraction}
Given a negation question $q$ at inference time, \textbf{SSP} transforms it into a structure-preserving skeleton $s$. The goal of skeleton extraction is to preserve the negation-aware question format while removing instance-specific semantic content. We first apply POS tagging and dependency parsing to identify the interrogative word, such as \texttt{what}, \texttt{which}, \texttt{where}, or \texttt{how}, and the negation operator, such as \texttt{not}, \texttt{no}, \texttt{without}. These tokens are preserved because they determine the question intent and the negated reasoning constraint.

After locating the interrogative word and the negation operator, \textbf{SSP} replaces the semantic span between them with a placeholder \texttt{X}. It then replaces the word or phrase immediately following the negation operator with a placeholder \texttt{Y}. If there is no semantic span between the interrogative word and the negation operator, the word following the negation operator is instead replaced with \texttt{X}. Formally, this process removes question-specific objects, attributes, and actions while retaining the key interrogative and negation structure.

For example, the question \texttt{[ Which animal is not standing? ]} is converted into the skeleton \texttt{[ Which X is not Y? ],} where \texttt{X} abstracts the queried visual entity and \texttt{Y} abstracts the negated action. The resulting skeleton serves as a question-type identifier and enables \textbf{SSP} to reuse reasoning strategies across questions with the same negation format as inference proceeds, reducing computational overhead. The algorithm of skeleton extraction is presented in Algorithm~\ref{alg:skeleton_extraction}.

\begin{algorithm}[t]
\caption{Question Skeleton Extraction}
\label{alg:skeleton_extraction}
\small
\begin{algorithmic}

\Require Negation question $q$
\Ensure Question skeleton $s$

\State Tokenize $q$ and obtain the part-of-speech tag of each word
\State Find the interrogative word or phrase $w$ among the first two words
\Comment{e.g., \texttt{what}, \texttt{where}, \texttt{which}, \texttt{do}, or \texttt{is}}

\State Find the negation word $n$ in $q$
\Comment{e.g., \texttt{not}, \texttt{no}, or \texttt{without}}

\State $B \gets$ the words located between $w$ and $n$

\If{$B$ contains a noun or noun phrase}
    \State Replace the noun or noun phrase in $B$ with \texttt{X}
\EndIf

\State Preserve any auxiliary or \textit{be} verb in $B$
\Comment{e.g., \texttt{is}, \texttt{are}, \texttt{do}, or \texttt{does}}

\State $A \gets$ the noun word following $n$

\If{$A$ exists}
    \State Replace $A$ with \texttt{Y}
\EndIf

\State $s \gets$ the resulting question
\State \Return $s$

\end{algorithmic}
\end{algorithm}

\subsection{Strategy Generation and Caching}
\label{StrategyGeneration}
\textbf{SSP} maintains an example pool $\mathcal{D}$ that maps each skeleton with a corresponding strategy, $\mathcal{D}[s] =  r,$ where $s$ is the skeleton extracted from the question and $r$ is the associated reasoning strategy.

For a test question $q$, \textbf{SSP} first extracts its skeleton $s$. If $s \in \mathcal{D}$, the corresponding cached $r$ is retrieved directly. Otherwise, \textbf{SSP} samples $N$ negative VQA pairs from a negative example pool. These examples are selected to share a similar negation-aware structure with the current question skeleton. The sampled examples are then provided to the VLM, which generates a concise one-sentence reasoning strategy describing how questions of this type should be solved. The generated strategy $r$ is stored together with the skeleton $s$ in $\mathcal{D}$ for future reuse. The prompt to generate the strategy is presented in the Appendix~\ref{App:strategy extraction}.

This caching mechanism allows \textbf{SSP} to convert recurring question formats into reusable prompting units. As more question types are observed during inference, the example pool gradually accumulates skeleton-strategy pairs, reducing redundant strategy generation and improving inference efficiency.

\begin{figure*}[t]
    \centering
    \includegraphics[width=0.9\linewidth]{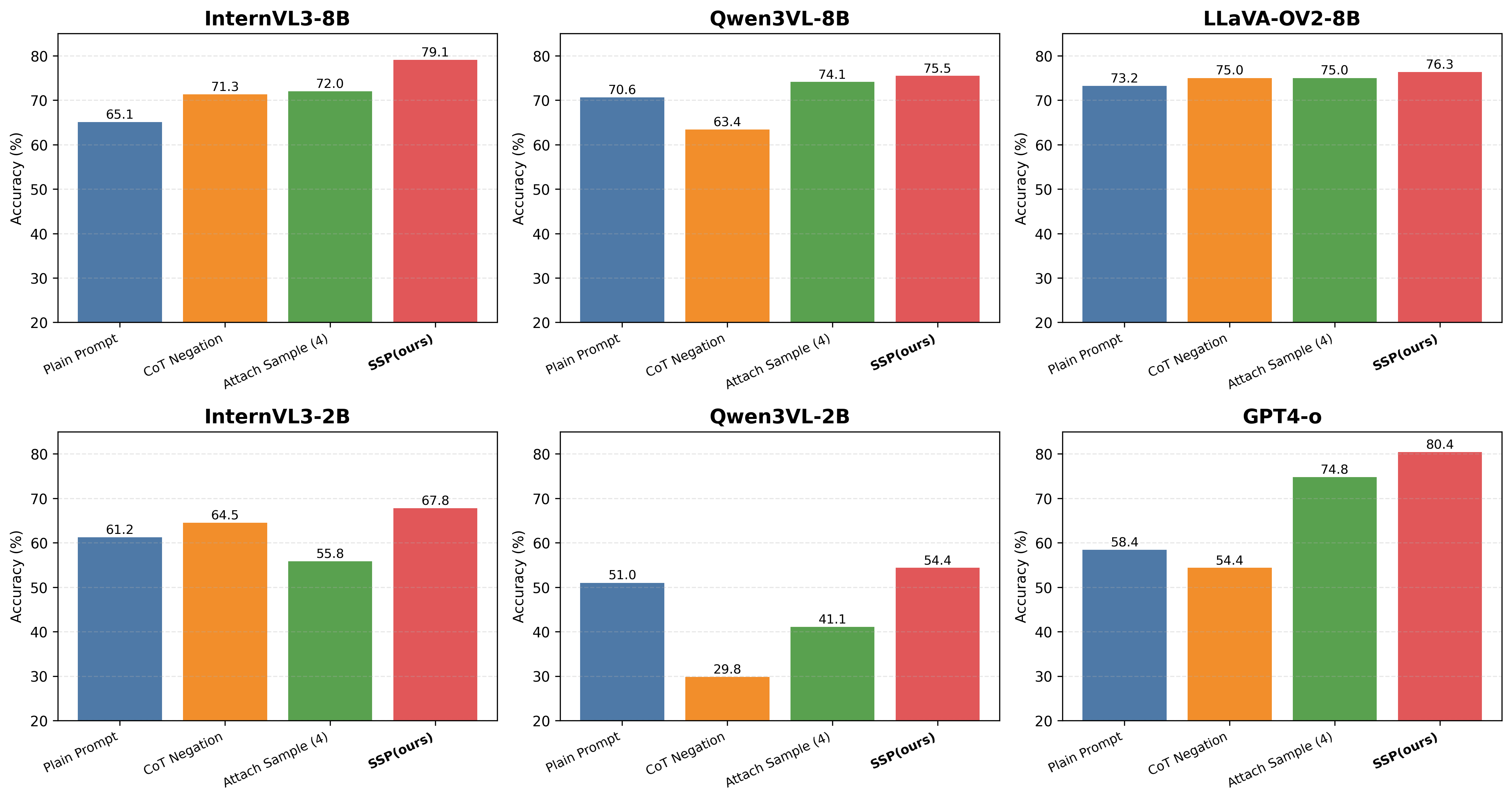}
    \caption{\textbf{Comparison of prompting methods on the NegVQA benchmark.} Accuracy (\%) is reported for six VLMs spanning commercial, 8B-level, and 2B-level models. \textbf{SSP} achieves the highest accuracy for every model, outperforming plain prompt, CoT negation and four-example in-context prompting, providing consistent gains across all evaluated model scales. }
    
    \label{fig:negvqa_comparison}
\end{figure*}

\begin{figure*}[t]
    \centering
    \includegraphics[width=0.9\linewidth]{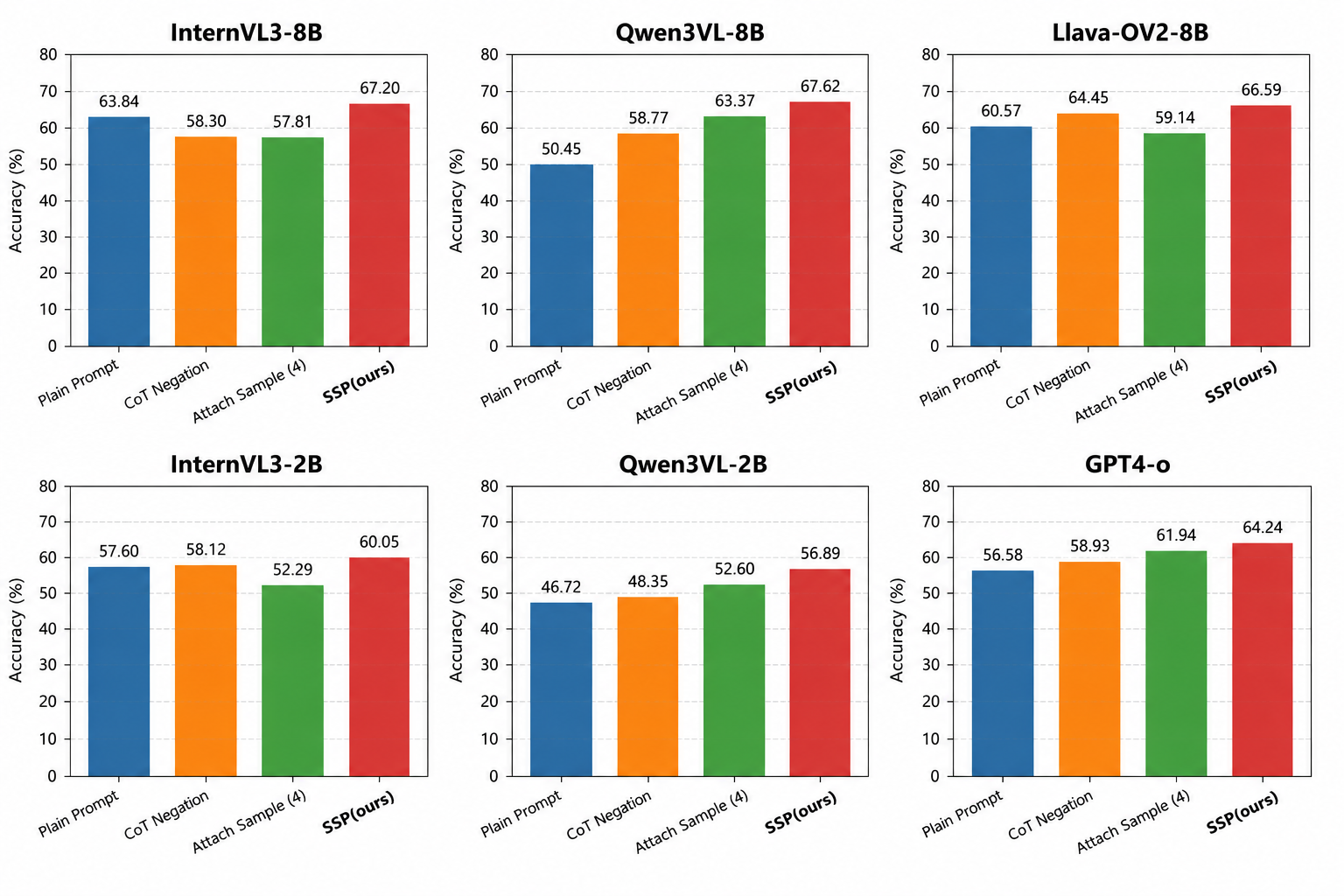}
    \vspace{-12pt}
    \caption{\textbf{Comparison of prompting methods on the DNVQA benchmark.} DNVQA evaluates implicit negation expressed without explicit markers such as \textit{not}, \textit{no}, or \textit{without}. Accuracy (\%) is reported for the same six VLMs evaluated on DNVQA. \textbf{SSP} achieves the highest accuracy for every model. These consistent gains show that \textbf{SSP} remains effective when negation is conveyed through diverse semantic expressions rather than explicit lexical cues.}
    
    \label{fig:dnvqa_comparison}
\end{figure*}
\subsection{Skeleton and Strategy Prompting}
Given a test image $I$, question $q$, extracted skeleton $s$, and retrieved or generated strategy $r$, \textbf{SSP} constructs the final prompt as:
[
\texttt{ Question Type: } s \texttt{ }
]
[
\texttt{ Reasoning Strategy: } r \texttt{ }
]
followed by the original image-question pair $(I, q)$. The skeleton informs the VLM of the abstract negation-aware question format, while the strategy provides high-level reasoning guidance for answering questions of this type.

The VLM then predicts the final answer conditioned on both the original visual-question input and the reusable skeleton-strategy prompt. Conceptually, \textbf{SSP} encourages the model to focus on the negated visual condition rather than relying only on surface-level linguistic cues or explicit demonstrations at test time.

\section{Experiments}
To evaluate the performance of \textbf{SSP} on negation VQA, we evaluate the proposed Skeleton and Strategy Prompting (\textbf{SSP}) framework on two benchmarks.
First, we consider the \textbf{NegVQA}~\cite{zhang2025negvqavisionlanguagemodels} benchmark, designed to test the negation understanding ability of vision-language models. It spans 20 widely-used VQA datasets across general, reasoning, OCR, and document and chart domains. We report accuracy as the evaluation metric. 
Second, we contribute and evaluate on the \textbf{Diverse Negation VQA Benchmark (DNVQA)}. 
We designed the DNVQA benchmark to evaluate the VLM's negation understanding capacity on questions that do not explicitly state negation words like \textit{"not, no, without"}. We generated 1000 negation VQA pairs via LLM and report accuracy as the evaluation metric (details in Appendix~\ref{app:NDVQA construction}).

We conduct experiments on three scales of VLMs. To represent large-scale models, we evaluate \textbf{GPT-4o}~\citep{openai2024gpt4ocard}; for 8B-level models, we evaluate \textbf{InternVL3-8B}~\citep{zhu2025internvl3exploringadvancedtraining}, \textbf{Qwen3VL-8B}~\citep{bai2025qwen3vltechnicalreport}, and \textbf{LLaVA-OneVision2-8B}~\citep{an2026llavaonevision2nextgenerationperceptualintelligence}; and for 2B-level models, we evaluate \textbf{InternVL3-2B}~\citep{zhu2025internvl3exploringadvancedtraining} and \textbf{Qwen3VL-2B}~\citep{bai2025qwen3vltechnicalreport}. This diverse set of VLMs allows us to examine whether \textbf{SSP} is effective across both large and lightweight VLMs.

We compare our \textbf{SSP} with three prompting baselines. The first baseline is \textbf{Plain Prompt}, where the original negation question is directly given to the VLM without any additional guidance. The second baseline is \textbf{CoT Negation}, where we prepend a hand-written rule to the question:
[
\texttt{Rule: For each question, convert it into positive question by removing the negation word. Then answer the positive question. Finally, choose the alternative answer than the positive question's answer.}
]
This baseline tests whether an explicit chain-of-thought-style negation rule can improve negation reasoning. The third baseline is \textbf{Attach Sample to Prompt}, where four negation question-answer examples are directly prepended before the test question as in-context demonstrations. 

\subsection{SSP Outperforms Baselines on NegVQA and DNVQA}

Figure~\ref{fig:negvqa_comparison} shows the comparison results on the NegVQA benchmark. Overall, the proposed Skeleton + Strategy prompting method achieves the best performance across all evaluated models. Compared with Plain Prompt, \textbf{SSP} improves InternVL3-8B from 65.1 to 79.1, Qwen3VL-8B from 70.6 to 75.5, and LLaVA-OneVision2-8B from 73.2 to 76.3. Similar improvements are observed on the smaller 2B-level models, where \textbf{SSP} improves InternVL3-2B from 61.2 to 67.8 and Qwen3VL-2B from 51.0 to 54.4. The results suggest that directly asking VLMs negation questions is insufficient for robust negation understanding. Although these models have strong visual-language capabilities, they can still struggle to correctly handle negated visual conditions without additional guidance. 
Compared with CoT Negation, \textbf{SSP} also achieves more stable performance. The hand-written CoT rule improves InternVL3-8B and InternVL3-2B, but it hurts Qwen3VL-8B and Qwen3VL-2B substantially. For example, Qwen3VL-2B drops from 51.0 with Plain Prompt to 29.8 with CoT Negation. 
A fixed rule for converting negation questions into positive questions may not generalize well across different model families.
Compared with directly attaching four negation examples to the prompt, \textbf{SSP} still performs better on all models. 
Simply adding a few negation examples is less effective than abstracting the question type and providing a reusable strategy. Rather than relying on instance-level demonstrations, \textbf{SSP} captures the common reasoning pattern behind a class of negation questions.

Meanwhile, \textbf{SSP} addresses this issue by explicitly providing an abstract question skeleton and a reusable reasoning strategy, allowing the model to focus on the negation-aware structure of the question. \textbf{SSP} consistently VLM model accuracy, showing that skeleton-guided and strategy-guided prompting provides more reliable negation reasoning guidance than other baselines.

Figure~\ref{fig:dnvqa_comparison} reports the results on DNVQA, which evaluates implicit negation expressed without the explicit markers \textit{not}, \textit{no}, or \textit{without}. \textbf{SSP} achieves the highest accuracy for all six evaluated models. Relative to Plain Prompt, it improves InternVL3-8B from 63.84 to 67.20, Qwen3VL-8B from 50.45 to 67.62, and LLaVA-OneVision2-8B from 60.57 to 66.59. The same pattern holds for the smaller models: accuracy increases from 57.60 to 60.05 for InternVL3-2B and from 46.72 to 56.89 for Qwen3VL-2B. \textbf{SSP} also improves GPT-4o from 56.58 to 64.24. These gains range from 2.45 to 17.17 percentage points, with the largest improvements observed for the Qwen3VL models.

The two prompting baselines behave less consistently on DNVQA. CoT Negation improves five models over Plain Prompt, but reduces InternVL3-8B accuracy from 63.84 to 58.30. Attaching four examples benefits Qwen3VL-8B, Qwen3VL-2B, and GPT-4o, but reduces accuracy for both InternVL3 models and LLaVA-OneVision2-8B. In contrast, \textbf{SSP} outperforms the strongest baseline for every model, with margins ranging from 1.93 points on InternVL3-2B to 4.29 points on Qwen3VL-2B. Together with the NegVQA results, these findings show that the effectiveness of \textbf{SSP} is not restricted to questions containing explicit negation words. Skeleton-conditioned strategies also improve performance when exclusion or absence is conveyed through more varied implicit expressions.

\subsection{SSP Components are Complementary}

\begin{table*}[t]
\centering
\small
\caption{Ablation study on 8B-level VLMs using the NegVQA benchmark. \textbf{SSP} achieves the best average performance, showing that combining matched skeletons with reusable strategies is more effective than using either component alone or using randomly selected skeleton-strategy pairs.}

\resizebox{\textwidth}{!}{
\begin{tabular}{lcccc}
\toprule
\textbf{Method} & \textbf{InternVL3-8B} & \textbf{Qwen3VL-8B} & \textbf{LLaVA-OV2-8B} & \textbf{Avg.} \\
\midrule
Skeleton Only & 72.3 & 70.0 & 73.7 & 72.0 \\
Strategy Only & 77.6 & 72.4 & 75.5 & 75.2 \\
Random Skeleton + Strategy & 76.1 & 71.0 & 75.4 & 74.2 \\
Skeleton + Strategy, one per sample & 78.7 & \textbf{75.8} & \textbf{76.3} & 76.9 \\
\textbf{SSP} & \textbf{79.1} & 75.5 & \textbf{76.3} & \textbf{77.0} \\
\bottomrule
\end{tabular}
}
\label{tab:ablation_8b}
\end{table*}

\begin{table}[t]
\centering
\small
\caption{\textbf{Efficiency of reusable strategy pairs on 1,000 NegVQA samples.}
Total strategy-generation time and generated-token count are reported when SSP reuses cached strategy pairs and when it generates a new strategy for every sample. Lower values indicate greater efficiency.}
\label{tab:strategy_reuse_efficiency}

\begin{tabular}{@{}lrrrr@{}}
\toprule
& \multicolumn{2}{c}{\textbf{Reusable strategy}}
& \multicolumn{2}{c}{\textbf{Per-sample strategy}} \\
\cmidrule(lr){2-3}
\cmidrule(l){4-5}
\textbf{Model}
& \textbf{Time (s)}
& \textbf{Tokens}
& \textbf{Time (s)}
& \textbf{Tokens} \\
\midrule
GPT-4o      & 51.45 & 489 & 2,237.84 & 20,317 \\
Qwen3VL-8B  & 35.97 & 463 & 1,564.76 & 20,114 \\
Qwen3VL-2B  & 27.19 & 423 & 1,182.17 & 18,914 \\
\bottomrule
\end{tabular}
\end{table}

We further conduct ablation studies on the 8B-level models to analyze the contribution of each component in \textbf{SSP}. We compare the full Skeleton + Strategy method with three variants: \textbf{Skeleton Only}, which removes the reasoning strategy and only provides the extracted skeleton; \textbf{Strategy Only}, which removes the skeleton and only provides the generated reasoning strategy; and \textbf{Random Skeleton + Strategy}, which randomly selects a skeleton-strategy pair instead of using the one matched to the current question type.

The ablation results show that both skeleton and strategy contribute to the final performance. Using only the skeleton achieves 72.3 on InternVL3-8B, 70.0 on Qwen3VL-8B, and 73.7 on LLaVA-OneVision2-8B, which is weaker than the full \textbf{SSP} method. 
The skeleton alone can provide useful structural information, but it does not fully specify how the model should reason over the negated visual condition.
Using only the strategy performs better than Skeleton Only, reaching 77.6 on InternVL3-8B, 72.4 on Qwen3VL-8B, and 75.5 on LLaVA-OneVision2-8B. 
The reasoning strategy is an important component of \textbf{SSP}. However, Strategy Only still underperforms the full method on InternVL3-8B and Qwen3VL-8B, showing that the skeleton provides complementary information by explicitly identifying the question type.

The Random Skeleton + Strategy variant also performs worse than the full method. For example, it obtains 76.1 on InternVL3-8B and 71.0 on Qwen3VL-8B, compared with 79.1 and 75.5 from the matched Skeleton + Strategy setting. This demonstrates that the gain of \textbf{SSP} does not simply come from adding any abstract instruction to the prompt. Instead, the skeleton and strategy need to match the current negation question type in order to provide effective guidance.

We also compare the cached \textbf{SSP} design with a variant that generates a new skeleton--strategy pair for every test sample. The per-sample variant achieves accuracies of 78.7, 75.8, and 76.3 on InternVL3-8B, Qwen3VL-8B, and LLaVA-OneVision2-8B, respectively. These results are comparable to those of cached \textbf{SSP}, which achieves 79.1, 75.5, and 76.3, respectively. Thus, reusing strategies changes accuracy by at most 0.4\% while substantially reducing strategy-generation overhead.
Table~\ref{tab:strategy_reuse_efficiency} quantifies this efficiency gain over 1,000 NegVQA samples. With reusable strategies, the three evaluated models require only 27.19--51.45 seconds and generate 423--489 output tokens. Generating a strategy separately for every sample instead requires 1,182.17--2,237.84 seconds and 18,914--20,317 output tokens. Strategy reuse therefore reduces generation time by approximately 43 times and generated-token count by approximately 42--45 times. These results show that the caching mechanism preserves the accuracy of per-sample strategy generation while making \textbf{SSP} substantially more efficient at inference time.

\subsection{Effective of Example Pool Size}

\begin{table*}[t]
\centering
\small
\begin{minipage}[t]{0.43\textwidth}
\centering
\begin{tabular}{cc}
\toprule
\textbf{Pool size} & \textbf{Accuracy (\%)} \\
\midrule
32  & 76.2 \\
64  & 76.8 \\
128 & 78.6 \\
256 & 79.1 \\
512 & 79.1 \\
\bottomrule
\end{tabular}
\end{minipage}
\hfill
\begin{minipage}[t]{0.43\textwidth}
\centering
\begin{tabular}{cc}
\toprule
\textbf{Number of samples} & \textbf{Accuracy (\%)} \\
\midrule
1 & 77.4 \\
2 & 78.3 \\
3 & 79.1 \\
4 & 79.1 \\
5 & 79.1 \\
\bottomrule
\end{tabular}
\end{minipage}

\caption{\textbf{Sensitivity of SSP to example-pool size and strategy-generation sample count on NegVQA using InternVL3-8B.}
\textbf{Left:} Accuracy increases as the example pool grows from 32 to 256 examples and saturates thereafter.
\textbf{Right:} Accuracy reaches its maximum when three examples are used to generate each strategy, with no further improvement from additional examples.}
\label{tab:pool_and_sample_analysis}
\end{table*}

\begin{table*}[t]
\centering
\small
\caption{Visualization of generated strategies for different negation-aware question skeletons. Each strategy is generated once for a skeleton and reused for later questions with the same question type.}

\begin{tabular}{p{0.35\textwidth}p{0.55\textwidth}}
\toprule
\textbf{Skeleton} & \textbf{Generated Strategy} \\
\midrule
What [X] is not [Y] & The correct answer to a negative question is the option that specifies the incorrect attribute or action. \\
How many [X] is not [Y] & Choose the option that represents the incorrect or less accurate number to the question.\\
Why [X] is not [Y] & The correct answer is that identifies the reason why the situation described in the question does not occur, which can be determined by selecting the option that describes an incorrect or unrelated circumstance. \\
Which [X] is not [Y] & Identifies the elements or elements that does not match the condition described in the question, and it can be determined by selecting the option that, if negated, would make the statement in the question true.\\
Is it not [X] & The correct answer is the option that contradicts the statement in the question, indicating the absence or incorrectness of the described condition. \\
\bottomrule
\end{tabular}
\label{tab}
\end{table*}

To further evaluate the the contribution of the example pool, we conduct experiments on variant example pool size, which contains 32, 64, 128, 256, and 512 negation VQA pairs respectively. The results are reported in Table~\ref{tab:pool_and_sample_analysis}. We observe an improvement in performance as the size of the example pool increases with small pool sizes. Specifically, increasing the pool size from 32 to 64 examples improves the accuracy from 76.2\% to 76.8\%, while further expanding the pool to 128 examples results in 78.6\%. The best performance of 79.1\% is achieved when the pool size is increased to 256 examples.
However, increasing the example pool beyond 256 examples does not provide additional benefits, as the performance remains unchanged when the pool size reaches 512 examples. This saturation phenomenon suggests that the proposed \textbf{SSP} framework can effectively capture the representative reasoning patterns of negation questions using a relatively compact example pool. Once sufficient diversity of negation structures and reasoning strategies has been covered, adding more examples introduces limited new information.

\subsection{Effect of Sample Number}

For each of the new question types, we retrieve 3 negation VQA pairs from the example pool. To evaluate the number of sampled pairs to the \textbf{SSP}'s performance, we conduct experiments with different number of sampled pair to generate different strategy for each question types. The results are reported in Table~\ref{tab:pool_and_sample_analysis}. 
Increasing the number of sampled examples generally improves the performance of the proposed \textbf{SSP} framework. When only one example is used to generate the negation instruction, the model achieves an accuracy of 77.4\%. Increasing the number of examples to two further improves the accuracy to 78.3\%, indicating that additional examples provide more comprehensive guidance for constructing robust reasoning strategies.
The performance reaches its maximum value of 79.1\% when three examples are used. Interestingly, further increasing the number of sampled examples to four or five does not yield additional improvements, suggesting that three representative examples are sufficient for capturing the essential reasoning patterns required for most negation question types.

\subsection{Generated Strategy Visualization}

To better understand how \textbf{SSP} guides the VLM, we visualize several generated strategies associated with different question skeletons. Since each strategy is generated at the skeleton level rather than the instance level, it captures the common reasoning pattern required by a group of negation questions, allowing us to examine whether the generated strategy correctly reflects the semantic role of the negation operator and whether it provides useful guidance for answering the corresponding question type.

Table~\ref{tab} shows examples of extracted skeletons and their corresponding generated strategies. Each skeleton abstracts away instance-specific objects, attributes, or actions using placeholders such as \texttt{X} and \texttt{Y}, while preserving the interrogative word and negation structure. The generated strategy then describes how the model should reason over questions with that skeleton. Ideally, a high-quality strategy should explicitly remind the model to focus on the negated condition, avoid answering the positive version of the question, and select the option or visual entity that satisfies the negative constraint.

\section{Conclusion}
We propose Skeleton-and-Strategy Prompting (\textbf{SSP}), a training-free inference-time framework for improving negation understanding in visual question answering. Rather than directly prompting VLMs with negation questions or relying only on instance-level demonstrations, \textbf{SSP} abstracts each question into a reusable negation-aware skeleton and associates it with a concise reasoning strategy. By caching skeleton-strategy pairs according to question type, \textbf{SSP} avoids repeated strategy generation and provides lightweight guidance that can be reused across test samples.
Experiments on the NegVQA benchmark and DNVQA benchmark show that \textbf{SSP} consistently improves performance across VLMs of different scales, including large, 8B-level, and 2B-level models. Ablation studies further demonstrate that both the skeleton and strategy components contribute to the final performance, while analyses on example pool size and sample number show that \textbf{SSP} can achieve strong results with a compact set of negative VQA examples. These findings suggest that explicit structure-aware prompting is an effective and efficient direction for improving VLM robustness on negation-based visual reasoning tasks.

\bibliography{main}
\bibliographystyle{tmlr}

\appendix
\section{Appendix}
\subsection{Strategy Extraction Pipeline}
\label{App:strategy extraction}
In Sec~\ref{StrategyGeneration}, we mentioned that if $t\notin \mathcal{D}$, we select $K$ negative VQA pairs from the negative example pool, which share the same skeleton. We then prompt the VLM with those samples to generate a concise one-sentence reasoning strategy description. Given a negative question $q$ with the image $I$, correct answer $a_t$, and wrong answer $a_f$, we first construct a uni-sample prompt with the specific format below:

\begin{tcolorbox}[
    colback=gray!5,
    colframe=gray!50,
    boxrule=0.5pt,
    arc=1pt,
    left=5pt,
    right=5pt,
    top=4pt,
    bottom=4pt
]
\small
\textbf{Example $i$:} \\
\textbf{Image:} $\langle \mathrm{image} \rangle$ \\
\textbf{Negative question:} $q$ \\
\textbf{Answer:} $a_t$, not $a_f$.
\end{tcolorbox}
We then combine the 4 uni-sample prompts with instruction as the prompt to generate the strategy via VLM:
```latex
\begin{tcolorbox}[
    colback=gray!5,
    colframe=gray!55,
    boxrule=0.5pt,
    arc=1pt,
    left=6pt,
    right=6pt,
    top=6pt,
    bottom=6pt
]
\small

You are given several visual question answering examples that all share the same question structure. 
For each, the image, the question, and the correct answer are provided.

\vspace{0.5em}

\textbf{Example 1:} \\
\textbf{Image:} $\langle \mathrm{image}_1 \rangle$ \\
\textbf{Negative question:} $q_1$ \\
\textbf{Answer:} $a_{t,1}$, not $a_{f,1}$.

\vspace{0.5em}

\textbf{Example 2:} \\
\textbf{Image:} $\langle \mathrm{image}_2 \rangle$ \\
\textbf{Negative question:} $q_2$ \\
\textbf{Answer:} $a_{t,2}$, not $a_{f,2}$.

\vspace{0.5em}

\textbf{Example 3:} \\
\textbf{Image:} $\langle \mathrm{image}_3 \rangle$ \\
\textbf{Negative question:} $q_3$ \\
\textbf{Answer:} $a_{t,3}$, not $a_{f,3}$.

\vspace{0.5em}

\textbf{Example 4:} \\
\textbf{Image:} $\langle \mathrm{image}_4 \rangle$ \\
\textbf{Negative question:} $q_4$ \\
\textbf{Answer:} $a_{t,4}$, not $a_{f,4}$.

\vspace{0.5em}

Provide a rule for determining the correct answer to negative questions in visual question answering. 
The rule should specify whether the correct answer is the positive or the incorrect option, and how to identify it, and must \textbf{NOT} mention the specific objects or scenes in the examples. 
Output it in one sentence beginning with ``RULE:''

\end{tcolorbox}

\subsection{DNVQA Benchmark Construction}
\label{app:NDVQA construction}
In addition to explicit negation questions that contain lexical negation markers such as \textit{not}, \textit{no}, or \textit{without}, we construct an diverse negation VQA benchmark (DNVQA), which contains the implicit negation question to evaluate whether VLMs can handle negation when it is expressed implicitly. Implicit negation refers to the questions whose semantic intent is negative, but whose surface form does not directly include explicit negation words. For example, instead of asking "\textit{Which object is not on the table?}", a soft negation question may ask "\textit{Which object is missing from the table?}" or "\textit{Which object is absent from the table?}" Such questions still require the model to reason about exclusion or absence, but the negation is conveyed through semantically negative expressions rather than explicit negation operators. 

To build the implicit negation data, we start from positive-form VQA examples consisting of an image, a question, the original correct answer, and a candidate answer set. The pipeline is shown in Algorithm~\ref{alg:soft_negation_construction}. For each example, we first identify the positive visual condition described by the question and answer. We then rewrite the question into an implicit negation form by replacing the positive intent with an implicit negative expression such as \textit{missing}, \textit{absent}, \textit{different from}, \textit{other than}, \textit{excluded}, or \textit{inconsistent with}. During rewriting, we avoid using explicit negation words, including \textit{not}, \textit{no}, \textit{never}, and \textit{without}. The target answer is then changed from the original positive answer to an alternative candidate that does not satisfy the original positive condition but satisfies the new soft negative question. This process produces examples that preserve the original image and answer space while changing the reasoning requirement from positive recognition to implicit negation understanding. We use an LLM-based rewriting step to generate fluent soft negation questions. To ensure data quality, each generated question is constrained to satisfy three requirements: (1) it must preserve the visual grounding of the original image, (2) it must express a negative or exclusion-based intent without explicit negation markers, and (3) its correct answer must be the alternative answer rather than the original positive answer. Examples that fail these constraints are discarded. The resulting soft negation data allows us to evaluate whether VLMs can recognize negation-like reasoning patterns beyond simple lexical cues. 

\begin{algorithm}[t] \caption{Soft Negation Data Construction} \label{alg:soft_negation_construction} \small \begin{algorithmic}[1] 
\Require Positive VQA dataset $\mathcal{D}$ \Require Explicit negation word set $\mathcal{N}$ \Ensure diverse negation dataset $\mathcal{D}_{\mathrm{implicit}}$ \State $\mathcal{D}_{\mathrm{implicit}} \gets \emptyset$ \For{each $(I_i, q_i, a_i, \mathcal{A}_i) \in \mathcal{D}$} \State Select $a_i^{-} \in \mathcal{A}_i$ such that $a_i^{-} \neq a_i$ \State Rewrite $q_i$ into a soft negation question $q_i^{\mathrm{implicit}}$ using an LLM \If{$q_i^{\mathrm{implicit}}$ contains any word in $\mathcal{N}$} \State Discard this example \State \textbf{continue} \EndIf \If{$q_i^{\mathrm{soft}}$ is visually grounded and $a_i^{-}$ is correct} \State Add $(I_i, q_i^{\mathrm{implicit}}, a_i^{-}, a_i)$ to $\mathcal{D}_{\mathrm{implicit}}$ \Else \State Discard this example \EndIf \EndFor \State \Return $\mathcal{D}_{\mathrm{implicit}}$ \end{algorithmic} \end{algorithm}

\end{document}